\documentclass[conference]{IEEEtran}
\usepackage{cite}
\usepackage{amsmath,amssymb,amsfonts}
\usepackage{algorithmic}
\usepackage{graphicx}
\usepackage{textcomp}
\usepackage[dvipsnames]{xcolor}
\begin{document}

\title{An Intelligent Assistant for Converting City Requirements to Formal Specification}

\author{
	\IEEEauthorblockN{Zirong Chen\IEEEauthorrefmark{1}, Isaac Li\IEEEauthorrefmark{2}, Haoxiang Zhang\IEEEauthorrefmark{3}, Sarah Preum\IEEEauthorrefmark{4}, John A. Stankovic\IEEEauthorrefmark{2}, Meiyi Ma\IEEEauthorrefmark{1}  \\}
	\IEEEauthorblockA{\IEEEauthorrefmark{1}
 Vanderbilt University  \IEEEauthorrefmark{2} University of Virginia
 \IEEEauthorrefmark{3}  Columbia University
 \IEEEauthorrefmark{4} Dartmouth College
 \\ \{zirong.chen, meiyi.ma\}@vanderbilt.edu \{il5fq, stankovic\}@virginia.edu sarah.masud.preum@dartmouth.edu
		}
	
}

\maketitle

\begin{abstract}

As more and more monitoring systems have been deployed to smart cities, there comes a higher demand for converting new human-specified requirements to machine-understandable formal specifications automatically. However, these human-specific requirements are often written in English and bring missing, inaccurate, or ambiguous information. In this paper, we present CitySpec~\cite{chen2022cityspec}, an intelligent assistant system for requirement specification in smart cities. CitySpec not only helps overcome the language differences brought by English requirements and formal specifications, but also offers solutions to those missing, inaccurate, or ambiguous information. The goal of this paper is to demonstrate how CitySpec works. Specifically, we present three demos: (1) interactive completion of requirements in CitySpec; (2) human-in-the-loop correction while CitySepc encounters exceptions; (3) online learning in CitySpec.
\end{abstract}


\section{Introduction}
\label{sec:introduction}

The demand for safety guarantees in smart cities has been increasing in recent years. Although monitoring systems have been developed and tested to be effective and efficient while dealing with machine-understandable formal specifications \cite{ma2021novel, ma2018cityresolver, ma2021predictive}, they require two important inputs, i.e., the real-time data streams and formal specified requirements. In other words, the inputs to the monitoring systems need to be machine-understandable formal specification streams. However, the city requirements given by the city policy makers are often written in English instead. The language difference between English specified requirements and their corresponding formal specifications is significant. The introduction of mathematical symbols even makes the translation more difficult. A case study of our dataset indicates that 2

Other than the language difference, the English requirements also introduce missing, inaccurate, or ambiguous information. In our dataset, there are lots of requirements that have missing parts, i.e., 27.6\% of the requirements lack location information, 29.1\% of the requirements miss a proper quantifier, and even 90\% of the requirements do not have or only have a default time (e.g., always) defined. Beside missing information, ambiguity and inaccuracy are found in the dataset as well, i.e., a specific number is needed to measure ``close to''.  As a result, it is considered difficult or impossible for monitoring systems to monitor these requirements properly.

In this demonstration, we show CitySpec \cite{chen2022cityspec}, an intelligent assistant system for smart city requirement specifications that bridges the gap between English requirements and formal specifications while also efficiently dealing with missing, inaccurate, or ambiguous information. CitySpec is designed to assist city policy makers in precisely filling out English city requirements using an intelligent interface, and then automatically converting them to formal specifications. The detailed system structure is introduced in Section \ref{sec:overview}. CitySpec is deployed online, Section \ref{sec:webpage} introduces the workflow of the online CitySpec tool. To evaluate the proposed system, we demonstrate three cases: (1) interactive completion of requirements in CitySpec; (2) human-in-the-loop correction while CitySepc encounters exceptions; (3) online learning in CitySpec in Section \ref{sec:evaluation}.
\section{CitySpec Overview}
\label{sec:overview}

CitySpec aims to help city policy makers write English requirements in an interactive interface and then coverts the collected requirements to formal specifications automatically. The system is deployed online.
CitySpec consists of four components:
\begin{itemize}
    \item An intelligent and interactive conversation-based \textit{interface} that helps communicate with city policy makers. Users first input a requirement in English, CitySpec then passes the requirement to the translation model and gets a formal requirement with keywords including, $\mathsf{entity}$: the requirement's main object, e.g., ``the number'', $\mathsf{quantifier}$: the scope of an entity, e.g., ``taxi'', $\mathsf{location}$: the location where this requirement is in effect, which is missing from the above example requirement, $\mathsf{time}$: the time period during which this requirement is in effect, e.g., ``between 7 am to 8 am'', $\mathsf{condition}$: the specific constraint on the entity, such as an upper or lower bound of $\mathsf{entity}$, e.g., ``10''. 
    \item A \textit{Requirement Synthesis} component that extracts city knowledge and synthesizes new requirements to build the translation model. The knowledge from city requirements is abstracted into vocabulary and pattern. Vocabulary stores semantic information and pattern stores syntactic information. To deal with data scarcity, a controllable synthesizing approach is introduced into this component. 
    \item A \textit{Translation Model} that converts city requirements to formal specifications. The inputs of the translation model are requirements, and the outputs of this component are formal specifications with token-level classification. We pick BiLSTM+CRF and BERT as the final reasoning neural models to handle the downstream NER task. Other NLP packages are applied to refine the final outputs, i.e., time refinement and negation detection.
    \item A \textit{Online Learning} component that adapts the system to new knowledge. When CitySpec encounters unknown/uncertain inputs, it will launch an online learning session to ask the user for clarification. Both short-term and long-term online learning methods are provided in this component. The short-term learning is designed to store the user clarification temporarily and accommodate the same user in one session of requirement specification with the temporary memory. Long-term learning is designed to adapt the new knowledge to the model permanently. To better filter out malicious or suspicious samples, a validation function is applied before the long-term learning.
\end{itemize}
\section{Webpage Workflow}
\label{sec:webpage}

CitySpec is deployed online and offers service with an online user interface. In this section, we introduce the workflow of the online CitySpec tool. There provides six example requirements for users to begin with at the bottom of our website:
\begin{itemize}
    \item The indoor concentrations of carbon monoxide should be no more than 7 mg/m3 in any 24 hours period in all the buildings.
    \item In all the buildings, annual average concentration of tetrachloroethylene should be no more than 0.25 mg/m3.
    \item All portable LED Luminaries should have Power Factor of no less than 0.70 everywhere.
    \item In all buildings, the average concentration of Bacterial should be no more than 2500 cfu/m3 for every day.
    \item The air quality within 3 miles of school A should always be better than moderate for the next 2 hours.
    \item The indoor concentrations of carbon monoxide should be no more than 7 mg/m3 in any 24 hours period in all the buildings.
\end{itemize}

\begin{figure}[h]
    \centering
    \includegraphics[width=1.1\columnwidth]{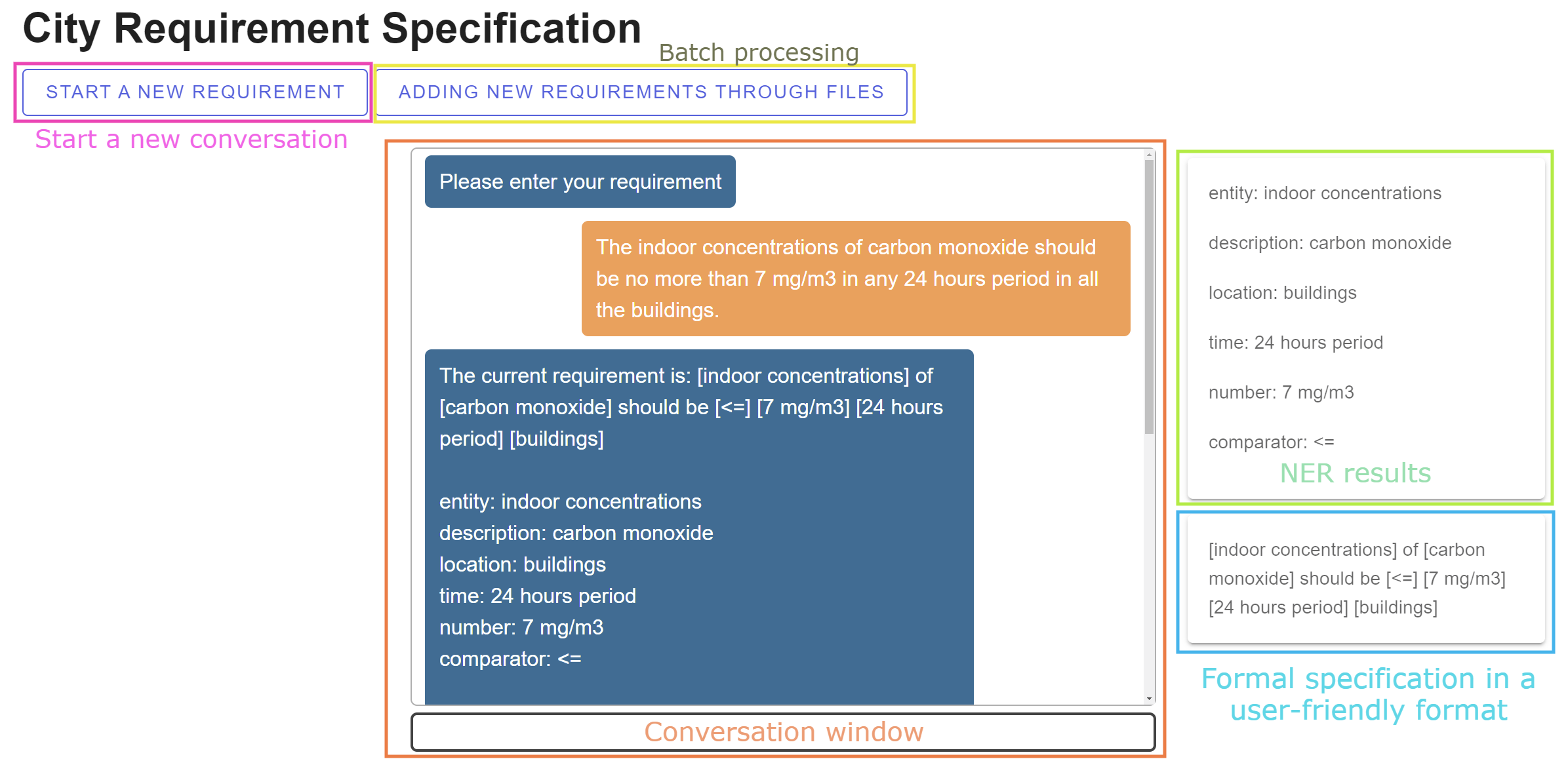}
    \caption{Online CitySpec tool layout}
    \label{fig:intro}
\end{figure}

As shown in Fig. \ref{fig:intro}, there are two buttons, two display windows, and one conversation window in the online CitySpec tool. In addition to the back-and-forth conversation on a single requirement specification, CitySpec also supports multiple requirement processing from the uploaded file. The top-left  {\textbf{start a new requirement}} button starts a new conversation while single requirement is given. The {\textbf{upload}} button on the right of the start button directs to a new page that supports file uploading and batch processing. The main body is the {\textbf{conversation window}} where the user can type in requirements, one requirement has already been typed in Fig. \ref{fig:intro}. On the right, there are two display windows, one for {\textbf{keyword results}}, another for generated \textbf{{formal specification}}. Keyword results are necessary references to generate final specifications. The window to display formal specifications modifies the final output a little bit. We consider our users, who are mainly city policy makers, do not have much expertise in formal languages. Thus, directly showing them the formal specifications can cause confusion and misunderstandings while users are trying to make corrections.

\section{Demonstration}
\label{sec:evaluation}

We demonstrate three use cases of CitySpec in this section: (1) interactive completion of requirements in CitySpec; (2) human-in-the-loop correction while CitySepc encounters exceptions; (3) online learning in CitySpec.

\begin{figure}[h]
    \centering
    \includegraphics[width=\columnwidth]{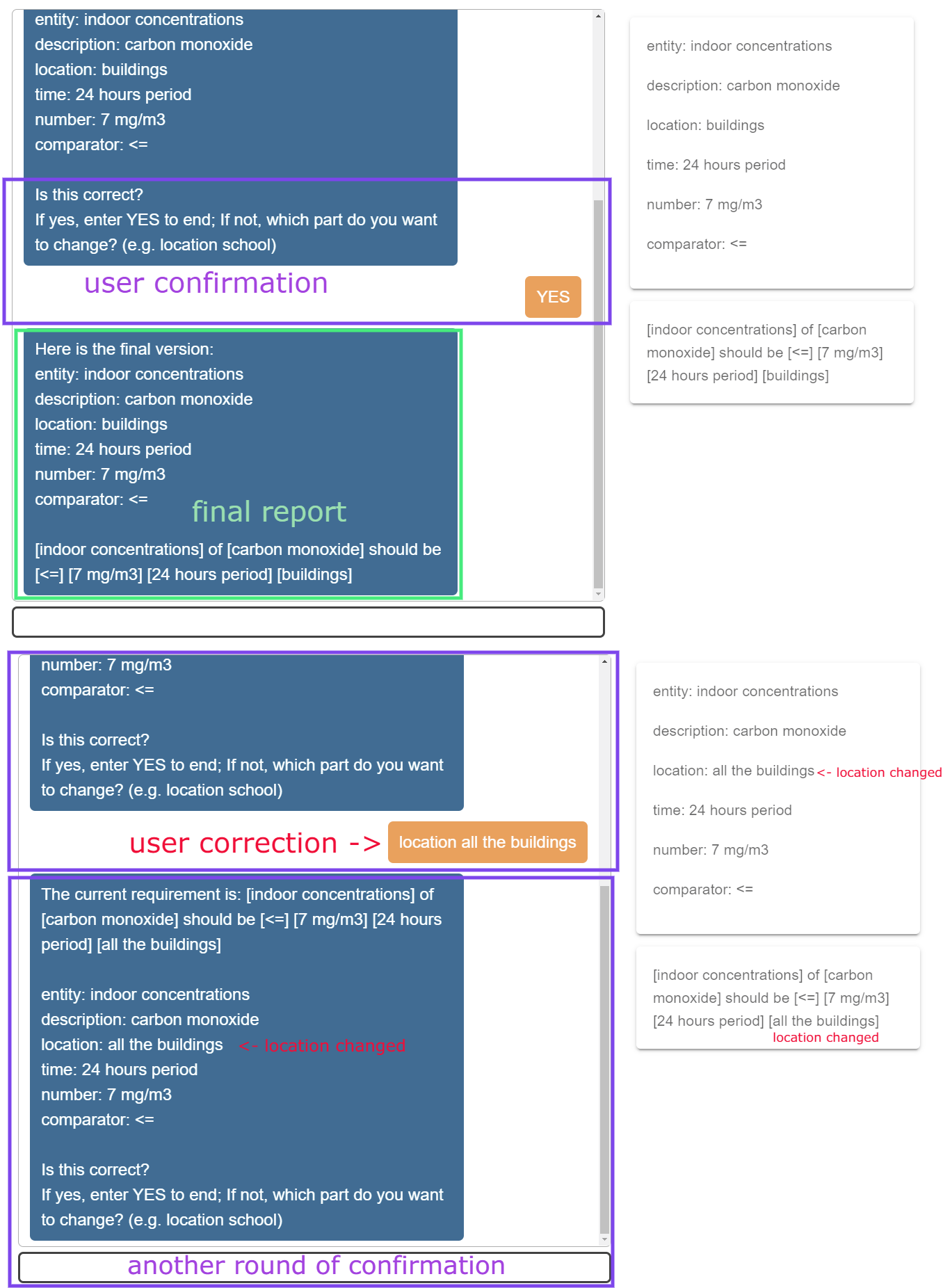}
    \caption{Case I (top) \& Case II (bottom) demonstration}
    \label{fig:case12}
\end{figure}

\subsection{Interactive Completion}
\label{subsec:IC}
When the user gives well-specified requirements, like `The indoor concentrations of carbon monoxide should be no more than 7 mg/m3 in any 24 hours period in all the buildings.' in Fig. \ref{fig:intro}, CitySpec calls its backend NER model to give intermediate NER results to fill in all the necessary domains for final formal specification generation. In the example shown in Fig. \ref{fig:intro}, $\mathsf{entity}$ is extracted as `indoor concentrations', $\mathsf{description}$ (aka $\mathsf{quantifier}$) is now `carbon monoxide', $\mathsf{location}$ is `buildings', $\mathsf{time}$ is `24 hours period', and $\mathsf{time}$ is `7 mg/m3'. After having all these preliminary NER results, CitySpec will throw a request for user confirmation before reporting the final results. If the user confirms, then the final results will be reported through both the conversation window and the two display windows on the right, see case I in Fig. \ref{fig:case12} for more details.

\subsection{Human-in-the-loop Correction}
\label{subsec:HC}
If the user does not confirm the preliminary results, then a human-in-the-loop correction will be needed, see case II in Fig. \ref{fig:case12}. Take the same requirement in Subsection \ref{subsec:IC} for example, if the user decides to use `all the buildings' as the $\mathsf{location}$ instead of `buildings', then all the user needs for correction is to type ``location all the buidlings'' in the conversation window to correct the preliminary results. After this, all preliminary $\mathsf{location}$ results will be changed as typed (see red arrows in Fig. \ref{fig:case12}). Similar to case I, no matter how many times the user changes the preliminary results, another round of user confirmation will be needed before CitySpec reports the final results.

\subsection{Short-term and Long-term Online Learning}

\begin{figure}[h]
    \centering
    \includegraphics[width=\columnwidth]{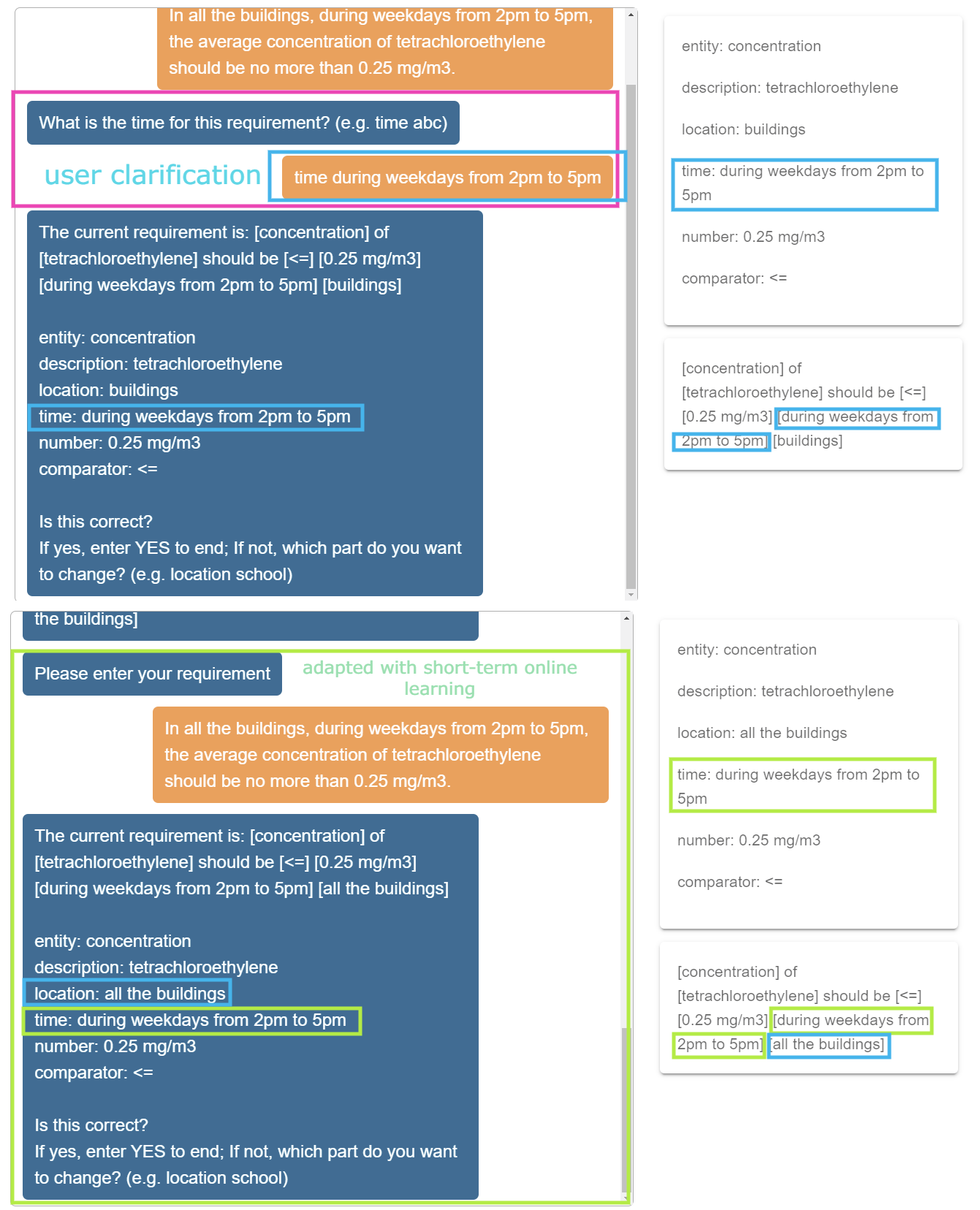}
    \caption{Case III demonstration}
    \label{fig:case3}
\end{figure}

CitySpec provides preliminary results in both cases in Subsection \ref{subsec:IC} and \ref{subsec:HC}. If CitySpec has trouble giving preliminary results, then human-in-the-loop correction and online learning will be launched. Take this requirement -- ``In all the buildings, during weekdays from 2pm to 5pm, the average concentration of tetrachloroethylene should be no more than 0.25 mg/m3.'' as an example, when CitySpec takes it as input, it has trouble giving $\mathsf{time}$ information, as a result, there is no preliminary result reported. Instead of asking for confirmation, CitySpec directly throws a request for user clarification about the $\mathsf{time}$ information. At the same time, if the user is not satisfied with the preliminary results, like the ``buildings'' in this case, the user can always change it halfway, see details in the first part of Fig. \ref{fig:case3}. After receiving clarification and correction, CitySpec stores it temporarily. When CitySpec is being queried with the same requirement, it will directly give the stored results, see the second image in Fig. \ref{fig:case3}. After the user signs out, the temporarily stored clarification is used for long-term online learning. Before injecting knowledge into the backend model, a validation function is applied to filter out malicious/suspicious inputs. After all, the deployed model is updated periodically with validated knowledge so that it promises to learn continuously.
\section{Conclusion}
In this demonstration abstract, we introduce CitySpec which is an intelligent assistant system for requirement specification, describe how it works in a component level, and demonstrate how it handles and further learn continuously from requirements with or without missing, inaccurate or ambiguous information.

\section*{Acknowledgment}
This work was funded, in part, by NSF CNS-1952096. 

\bibliographystyle{IEEEtran}
\bibliography{refs}

\begin{thebibliography}{1}
\providecommand{\url}[1]{#1}
\csname url@samestyle\endcsname
\providecommand{\newblock}{\relax}
\providecommand{\bibinfo}[2]{#2}
\providecommand{\BIBentrySTDinterwordspacing}{\spaceskip=0pt\relax}
\providecommand{\BIBentryALTinterwordstretchfactor}{4}
\providecommand{\BIBentryALTinterwordspacing}{\spaceskip=\fontdimen2\font plus
\BIBentryALTinterwordstretchfactor\fontdimen3\font minus
  \fontdimen4\font\relax}
\providecommand{\BIBforeignlanguage}[2]{{%
\expandafter\ifx\csname l@#1\endcsname\relax
\typeout{** WARNING: IEEEtran.bst: No hyphenation pattern has been}%
\typeout{** loaded for the language `#1'. Using the pattern for}%
\typeout{** the default language instead.}%
\else
\language=\csname l@#1\endcsname
\fi
#2}}
\providecommand{\BIBdecl}{\relax}
\BIBdecl

\bibitem{chen2022cityspec}
Z.~Chen, I.~Li, Z.~Haoxiang, S.~Preum, J.~A. Stankovic, and M.~Ma, ``Cityspec:
  An intelligent assistant system for requirement specification in smart
  cities,'' in \emph{2022 IEEE International Conference on Smart Computing
  (SMARTCOMP)}.\hskip 1em plus 0.5em minus 0.4em\relax IEEE, 2022.

\bibitem{ma2021novel}
M.~Ma, E.~Bartocci, E.~Lifland, J.~A. Stankovic, and L.~Feng, ``A novel
  spatial--temporal specification-based monitoring system for smart cities,''
  \emph{IEEE Internet of Things Journal}, vol.~8, no.~15, pp. 11\,793--11\,806,
  2021.

\bibitem{ma2018cityresolver}
M.~Ma, J.~A. Stankovic, and L.~Feng, ``Cityresolver: a decision support system
  for conflict resolution in smart cities,'' in \emph{2018 ACM/IEEE 9th
  International Conference on Cyber-Physical Systems (ICCPS)}.\hskip 1em plus
  0.5em minus 0.4em\relax IEEE, 2018, pp. 55--64.

\bibitem{ma2021predictive}
M.~Ma, J.~Stankovic, E.~Bartocci, and L.~Feng, ``Predictive monitoring with
  logic-calibrated uncertainty for cyber-physical systems,'' \emph{ACM
  Transactions on Embedded Computing Systems (TECS)}, vol.~20, no.~5s, pp.
  1--25, 2021.

\end{thebibliography}

\end{document}